\documentclass[conference]{IEEEtran}
\IEEEoverridecommandlockouts

\usepackage{cite}
\usepackage{amsmath,amssymb,amsfonts}
\usepackage{graphicx}
\usepackage{textcomp}
\usepackage{xcolor}
\usepackage{booktabs}
\usepackage{multirow}
\usepackage{url}
\usepackage{algorithm}
\usepackage{algpseudocode}
\usepackage{tikz}
\usepackage{pgfplots}
\usepackage{enumitem}

\newcommand{\frameworkname}{ST-ProC}

\begin{document}

\title{ST-ProC: A Graph-Prototypical Framework for Robust Semi-Supervised Travel Mode Identification}

\author{\IEEEauthorblockN{Luyao Niu}
\IEEEauthorblockA{\textit{Peking University} \\
Shenzhen, China \\
luyao0160@stu.pku.edu.cn}
\and
\IEEEauthorblockN{Nuoxian Huang}
\IEEEauthorblockA{\textit{Imperial College London} \\
London, UK \\
n.huang25@imperial.ac.uk}
}

\maketitle

\begin{abstract}
Travel mode identification (TMI) from GPS trajectories is critical for urban intelligence, but is hampered by the high cost of annotation, leading to severe label scarcity. Prevailing semi-supervised learning (SSL) methods are ill-suited for this task, as they suffer from catastrophic confirmation bias and ignore the intrinsic data manifold. We propose \frameworkname{}, a novel graph-prototypical multi-objective SSL framework to address these limitations. Our framework synergizes a graph-prototypical core with foundational SSL Support. The core exploits the data manifold via graph regularization, prototypical anchoring, and a novel, margin-aware pseudo-labeling strategy to actively reject noise. This core is supported and stabilized by foundational contrastive and teacher-student consistency losses, ensuring high-quality representations and robust optimization. \frameworkname{} outperforms all baselines by a significant margin, demonstrating its efficacy in real-world sparse-label settings, with a performance boost of 21.5\% over state-of-the-art methods like FixMatch.
\end{abstract}

\begin{IEEEkeywords}
Travel Mode Identification, Semi-Supervised Learning, Graph Regularization, Prototypical Networks, Contrastive Learning
\end{IEEEkeywords}

\section{Introduction}
\label{sec:introduction}
% P1: 意义和重要性 (Unchanged)
The proliferation of location-aware devices has led to an explosive growth in human trajectory data. A fundamental task in mobility analysis is Travel Mode Identification (TMI), which aims to classify travel modes (e.g., walk, bike, bus) from raw GPS data ~\cite{zeng2025advancing}. TMI serves as a cornerstone for high-impact applications, including intelligent transportation systems, carbon footprint estimation, personalized location-based services, and smart city planning.

% P2: 早期方法 -> RNNs (Unchanged)
Early TMI methods relied on traditional machine learning (e.g., SVMs, Random Forests) with extensive hand-crafted features \cite{gonzalez2010automating}. While insightful, these methods suffer from limited representation power. This led to the adoption of deep learning models, such as LSTMs, GRUs \cite{james2020travel}, and Transformers, which can automatically learn from sequential patterns.

% P3: GNNs (及所有DL方法) 的核心挑战：监督 (Unchanged)
Despite these advances, a dominant challenge remains. Most deep learning models require massive, accurately labeled datasets to perform well. However, acquiring such ground-truth labels is expensive and labor-intensive, typically requiring manual user annotation. This makes purely supervised approaches impractical for real-world, large-scale deployment \cite{zeng2025advancing}.

% P4: 引出SSL的必要性，以及现有SSL的致命缺陷 (Unchanged)
This label scarcity strongly motivates a shift towards SSL. However, prevailing general-purpose SSL frameworks (e.g., FixMatch \cite{sohn2020fixmatch}) are ill-suited for the sparse TMI task. First, they suffer from catastrophic confirmation bias in extreme label scarcity \cite{arazo2020pseudo}. When labels are scarce, these models progressively reinforce their own noisy pseudo-labels, especially on ambiguous modes with subtle motion differences, leading to degenerated solutions. Second, they  treat samples as independent and identically distributed (i.i.d.) and ignored the latent graph structure that governs the data. This i.i.d. assumption fails to leverage the fact that trajectories are not independent (e.g., segments on the same rail line share a label), thus missing a critical source of regularization \cite{yu2023graph}.

% P5: 抛出其他挑战，并引出我们的解决方案 (Unchanged)
Furthermore, many methods attempt to compensate for poor performance by introducing external data \cite{bantis2017you}, such as road networks or POIs. This reliance compromises generalizability and introduces privacy risks.

% P6: 我们的解决方案 (Polished and Rewritten)
% This paragraph is heavily refined to directly answer P4, avoiding AI-tells.
To address these specific limitations, we propose \frameworkname{}, a novel, context-free, graph-enhanced SSL framework. Our approach directly tackles the two SSL challenges. To combat confirmation bias, we introduce a graph-prototypical core. This core uses semantic prototypes, derived from labeled data, to act as stable anchors. It then employs a graph-based, margin-aware filter to actively reject noisy pseudo-labels, rather than passively accepting them. To exploit the latent graph structure, this core explicitly builds an endogenous semantic graph to enforce manifold consistency through regularization. We stabilize this entire process with foundational SSL Support (contrastive learning and teacher-student consistency) in a multi-objective design. This structure allows the model to learn robust representations while simultaneously respecting the data's intrinsic geometry and mitigating label noise.

% P7: 贡献 (Polished and Tightened)
In summary, our contributions are as follows:
\begin{enumerate}[leftmargin=*] 
    \item Propose \frameworkname{}, a novel multi-objective SSL framework for sparse-label TMI that integrates graph regularization and prototypical anchoring to explicitly model the underlying topological structure of trajectory data.
    \item Develop a robust, dual-filtered pseudo-labeling strategy that fuses graph and prototype predictions to actively mitigate confirmation bias in sparse regimes..
    \item Achieve state-of-the-art performance on the GeoLife benchmark, significantly outperforming strong baselines.
\end{enumerate}

% P8: 路线图 (Unchanged)
The remainder of this paper is organized as follows. Section \ref{sec:methodology} details our proposed framework. Section \ref{sec:experiments} presents the experimental setup and results, followed by a conclusion in Section \ref{sec:conclusion}.

\section{Methodology}
\label{sec:methodology}

\subsection{Problem Formulation}
We formulate TMI as a semi-supervised classification task on trajectory data. The complete dataset $\mathcal{D}$ is composed of a labeled subset $\mathcal{D}_L = \{(\mathcal{T}_i, y_i)\}_{i=1}^{N_L}$ (where $y_i \in \{1, \dots, K\}$ is the ground-truth mode) and a significantly larger unlabeled subset $\mathcal{D}_U = \{\mathcal{T}_i\}_{i=1}^{N_U}$. Each raw trajectory $\mathcal{T}_i = \{(t_j, lat_j, lon_j)\}_{j=1}^{L_i}$ is a temporal sequence of GPS coordinates. Our objective is to learn a robust mapping function $\Phi: \mathcal{T} \to \{1, \dots, K\}$ that accurately identifies transport modes. The core challenge is effectively generalizing from the sparse supervision in $\mathcal{D}_L$ by mining the latent structural information within $\mathcal{D}_U$, without relying on external geographic context.

\subsection{Framework Overview}
To tackle the inherent catastrophic confirmation bias and manifold ignorance in sparse-label TMI, we propose \frameworkname{}, a graph-prototypical multi-objective framework. As illustrated in Fig \ref{fig:framework}, the pipeline consists of three synergistic stages:

\begin{figure*}[h!]
    \centering
    \includegraphics[width=\textwidth]{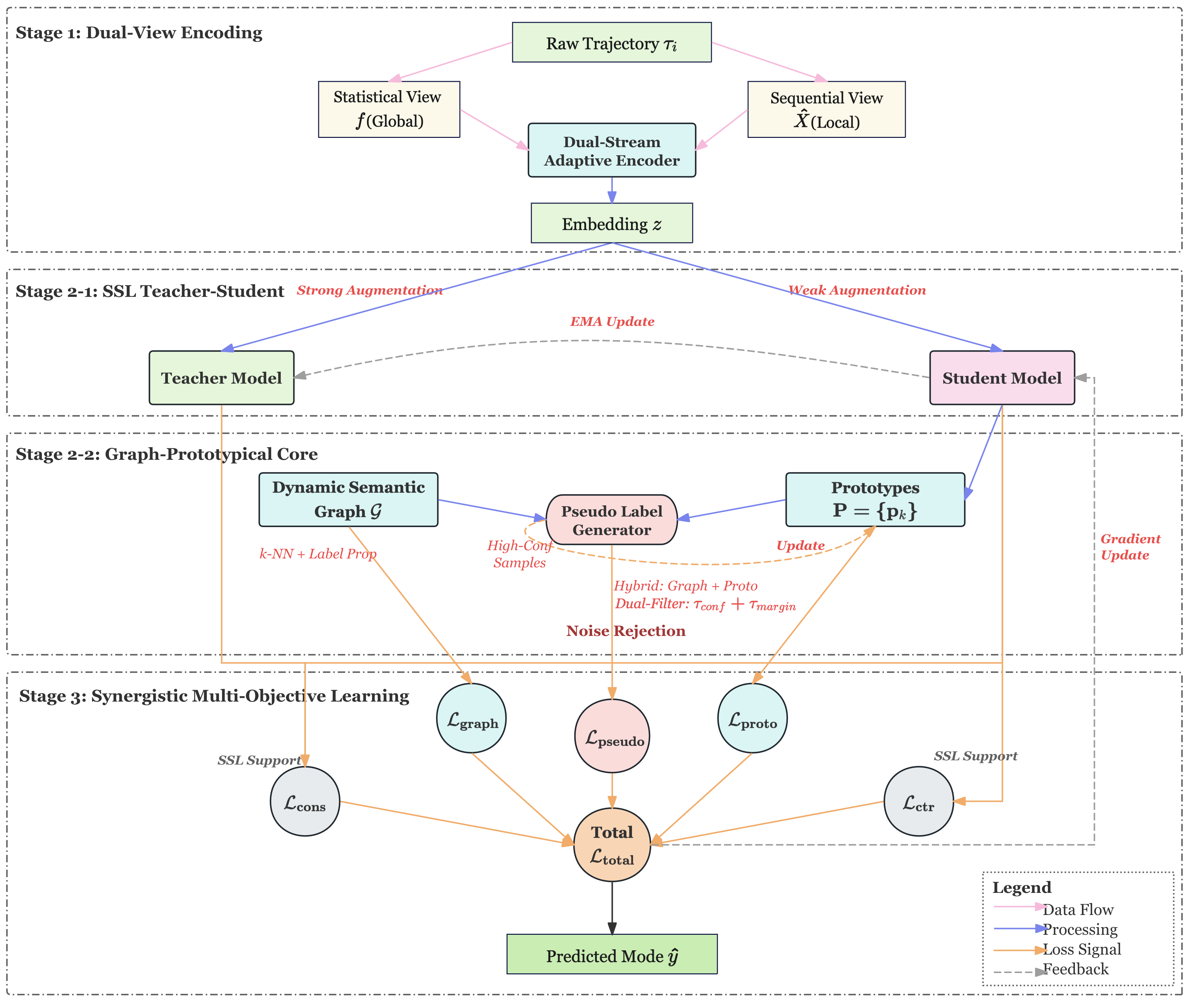} 
        \caption{\textbf{Schematic illustration of the \frameworkname{} framework.}}
    \label{fig:framework}
\end{figure*}

\begin{enumerate}[leftmargin=*] 
    \item \textbf{Dual-View Representation Encoding:} Raw trajectories are synthesized into complementary statistical and sequential views. An adaptive dual-stream encoder then fuses these heterogeneous features into a unified latent embedding $\mathbf{z}$.
    \item \textbf{Dynamic Semantic Graph Construction:} The data manifold is explicitly modeled by constructing a global semantic graph $\mathcal{G}$ over the embedded samples, capturing intrinsic relationships among data points.
    \item \textbf{Synergistic Multi-Objective Learning:} The encoder is optimized using a cohesive set of five objectives that strategically integrate (a) foundational contrastive learning, (b) prototypical anchoring, (c) manifold graph regularization, (d) confidence-aware pseudo-labeling, and (e) teacher-student consistency.
\end{enumerate}
Collectively, \frameworkname{} innovates by tightly coupling manifold-aware graph regularization and prototypical semantic anchoring within a stabilized multi-objective loss, enabling the model to effectively utilize the intrinsic structure of unlabeled data while actively suppressing catastrophic confirmation bias.

\subsection{Trajectory Representation and Encoder}
\label{sec:encoder}
To generate high-quality representations suitable for graph propagation, we engineer two complementary views from the raw trajectory $\mathcal{T}_i$ (projected to a 2D planar system). The first is a holistic statistical vector ($\mathbf{f}$), which is a dense feature summarizing global kinematics, geometric tortuosity, and stop-and-dwell characteristics. The second is a multi-channel motion sequence ($\tilde{\mathbf{X}}$), a 4-channel time-series $[x, y, v_x, v_y]$ capturing fine-grained, step-by-step dynamic patterns. 

These views are processed by an adaptive dual-stream encoder. A spatio-temporal stream (Transformer coupled with BiGRU) extracts sequential dependencies from $\tilde{\mathbf{X}}$, while a statistical stream (MLP) encodes global properties from $\mathbf{f}$. A learnable gating mechanism adaptively fuses these heterogeneous representations into a comprehensive hidden state $\mathbf{h}$. Last, a projection head maps $\mathbf{h}$ to a normalized $D$-dimensional embedding $\mathbf{z}$ ($||\mathbf{z}||=1$), which serves as the canonical input for our downstream semi-supervised framework.

\subsection{Graph-Prototypical Multi-Objective SSL Framework}
\label{sec:ssl_framework}
The efficacy of \frameworkname{} stems from the synergistic optimization of five objectives within a shared embedding space $\mathbf{z}$. This design ensures the learned representations are not only discriminative but also compliant with the underlying mobility manifold of GPS trajectories.

\subsubsection{Dynamic Semantic Graph Construction}
\label{sec:graph_construction}
Standard i.i.d. assumptions fail in TMI because trajectories are governed by physical infrastructure: tracks on the same road segment or rail line typically belong to the same transport mode (e.g., subway) regardless of individual speed variations. To capture this topological dependency, we explicitly model the data manifold. We periodically reconstruct a global $k$-NN graph $\mathcal{G}_{\text{global}} = (\mathcal{V}, \mathcal{E})$ over the embeddings $\mathbf{Z}$, using cosine similarity to link kinetically and spatially similar trajectories.
To inject this stable, global manifold structure into local training, we derive a batch-wise adjacency matrix $\mathbf{A}_b$ via subgraph clipping. This ensures that local optimization respects the global continuity of the transportation network. In the initial warmup phase, we fallback to an on-the-fly dynamic $k$-NN graph computed within the batch.

\subsubsection{Foundational Contrastive Learning}
Raw GPS data is inherently susceptible to sensor noise, sampling rate irregularities, and signal drift. To establish a latent space robust to these perturbations, we employ a foundational contrastive objective. Let $\mathbf{z}_1$ and $\mathbf{z}_2$ be embeddings of two augmented views of the same trajectory, where augmentations simulate realistic GPS anomalies (e.g., point masking or jittering). We minimize the NT-Xent loss:
\begin{equation}
    \mathcal{L}_{\text{ctr}} = -\mathbb{E}_{i} \left[ \log \frac{\exp(\mathbf{z}_{1,i} \cdot \mathbf{z}_{2,i} / \tau_c)}{\sum_{k=1}^{2B} \mathbb{1}_{k \neq i} \exp(\mathbf{z}_{1,i} \cdot \mathbf{z}_{k} / \tau_c)} \right]
\end{equation}

This objective pulls together views of the same underlying movement pattern, forcing the model to learn representations invariant to sensor-induced variations rather than overfitting to specific coordinate sequences.

\subsubsection{Prototypical Anchoring}
In sparse-label TMI, distinguishing modes with similar kinematics  is challenging. To mitigate semantic drift, we establish stable class representations by maintaining a set of learnable prototypes $\mathbf{P} = \{\mathbf{p}_k\}_{k=1}^K$, where each prototype represents the canonical motion signature of a transport mode (e.g., the stop-and-go pattern of a bus). The probability of sample $\mathbf{z}_i$ belonging to class $k$ is modeled via:
\begin{equation}
    p(y=k | \mathbf{z}_i) = \frac{\exp(\mathbf{z}_i^\top \mathbf{p}_k / \tau_p)}{\sum_{j=1}^K \exp(\mathbf{z}_i^\top \mathbf{p}_j / \tau_p)}
\end{equation}

For labeled samples, we minimize $\mathcal{L}_{\text{proto}}$ to anchor prototypes to true class semantics:
\begin{equation}
    \mathcal{L}_{\text{proto}} = -\mathbb{E}_{(\mathbf{z}_i, y_i) \in \mathcal{D}_L} \left[ \log p(y_i | \mathbf{z}_i) \right]
\end{equation}

Prototypes are updated via Exponential Moving Average (EMA) using high-confidence samples, ensuring they evolve stably even when labeled data covers only a fraction of the road network.

\subsubsection{Graph Regularization}
To enforce the manifold hypothesis—that trajectories traversing the same physical path should share semantic labels—we introduce a dual graph regularization mechanism based on $\mathbf{A}_b$:
\begin{enumerate}[leftmargin=*] 
    \item \textbf{Laplacian Smoothness ($\mathcal{L}_{\text{graph\_smooth}}$):} We utilize the graph Laplacian $\mathbf{L}_{\text{Lap}}$ to penalize abrupt semantic shifts between connected nodes (e.g., ensuring a trajectory segment on a railway is not isolatedly classified as car):
    \begin{equation}
        \mathcal{L}_{\text{graph\_smooth}} = \frac{1}{|\mathcal{B}|} \text{Tr}(\mathbf{Z}^\top \mathbf{L}_{\text{Lap}} \mathbf{Z})
    \end{equation}
    \item \textbf{Neighbor Contrast ($\mathcal{L}_{\text{nbr\_ctr}}$):} We further refine the local structure by treating neighbors $\mathcal{N}(i)$ (trajectories with similar motion contexts) as positive pairs:
    \begin{equation}
        \mathcal{L}_{\text{nbr\_ctr}} = -\mathbb{E}_{\mathbf{z}_i} \left[ \log \frac{\sum_{\mathbf{z}_j \in \mathcal{N}(i)} \exp(\mathbf{z}_i \cdot \mathbf{z}_j / \tau_n)}{\sum_{\mathbf{z}_k \neq \mathbf{z}_i} \exp(\mathbf{z}_i \cdot \mathbf{z}_k / \tau_n)} \right]
    \end{equation}
\end{enumerate}
These regularizers ensure the learned embedding space mirrors the topological continuity of the real-world transportation network.

\subsubsection{Robust Graph-based Pseudo-Labeling}
Ambiguous motion patterns (e.g., walking vs. waiting for a bus) often generate noisy predictions in TMI. This component serves as our primary defense against confirmation bias by expanding supervision to $\mathcal{D}_U$ via a rigorous filter.

Pseudo-labels $\hat{y}_i$ and confidence scores $c_i$ are generated by fusing predictions from (a) direct prototype similarity and (b) label propagation across the global graph $\mathcal{G}_{\text{global}}$, effectively smoothing out local noise.

A pseudo-label is accepted only if it satisfies both a \textit{high-confidence threshold} ($c_i > \tau_{\text{conf}}$) and a \textit{high-margin criterion} ($m_i = c_i^{(1)} - c_i^{(2)} > \tau_{\text{margin}}$), rejecting ambiguous samples that sit on decision boundaries (e.g., transition points between modes).

The loss is formulated as a confidence-weighted cross-entropy loss on the filtered subset $\mathcal{D}_{\text{filtered}}$:
\begin{equation}
    \mathcal{L}_{\text{pseudo}} = -\mathbb{E}_{\mathbf{z}_i \in \mathcal{D}_{\text{filtered}}} \left[ c_i \cdot \log p(\hat{y}_i | \mathbf{z}_i) \right]
\end{equation}

\subsubsection{Teacher-Student Consistency}
Last, to stabilize training against the stochastic nature of GPS augmentations, we incorporate a teacher-student  consistency mechanism. The Teacher model (EMA-updated weights $\theta_t$) provides stable targets, mitigating the variance caused by aggressive data augmentation (e.g., large-scale masking):
\begin{equation}
    \mathcal{L}_{\text{cons}} = \mathbb{E}_{\mathbf{z}} \left[ || \mathbf{z}_1 - \mathbf{z}_t ||_2^2 \right]
\end{equation}

This ensures that the model yields consistent predictions for a trajectory regardless of partial signal loss or noise injection.

\subsubsection{Optimization Strategy}
The model is trained end-to-end by minimizing a total objective that balances the five loss components. To prevent early overfitting to potentially noisy pseudo-labels, we employ time-dependent ramp-up functions, $w_p(t)$ and $w_c(t)$, for the semi-supervised terms:
\begin{equation}
\begin{aligned}
    \mathcal{L}_{\text{total}} = & \mathcal{L}_{\text{ctr}} + \lambda_p \mathcal{L}_{\text{proto}} \\
                         & + \lambda_s \mathcal{L}_{\text{graph\_smooth}} + \lambda_n \mathcal{L}_{\text{nbr\_ctr}} \\
                         & + w_p(t) \cdot \mathcal{L}_{\text{pseudo}} + w_c(t) \cdot \mathcal{L}_{\text{cons}}
\end{aligned}
\end{equation}

This curriculum-based optimization allows the embeddings and prototypes to mature before the model relies heavily on propagated labels and consistency constraints.

\section{Experiments}
\label{sec:experiments}
We first detail the experimental setup and training process, then provide a comparison experiment and detailed class analysis.

\subsection{Experimental Setup}
\label{sec:exp_setup}
We evaluate our method on the \textit{GeoLife} Trajectories dataset focusing on 5 main classes: walk, bike, bus, car (incl. taxi), and subway. We simulate the SSL environment by masking training labels across varied ratios, enforcing a minimum of 15 labeled samples per class for low ratios. A WeightedRandomSampler is used to balance the sampling of labeled and unlabeled data.

Our dual-stream encoder ingests 4-channel sequences and $F=48$ statistical features. We use the AdamW optimizer with a cosine annealing schedule (5-epoch warmup), early stopping, and radient clipping. The total loss weights are: $\lambda_p=1.0$, $\lambda_c=0.2$, $\lambda_s=0.05$, $\lambda_n=0.10$, and $\lambda_{pseudo}=0.5$. Key hyperparameters include $\tau=0.1$ (contrastive), $m=0.15$ (proto margin), $\alpha=0.999$ (EMA), and $k=10$ (graph $k$-NN).

\subsection{Baselines}
We compare \frameworkname{} against unsupervised, semi-Supervised, and supervised strong baselines. To ensure a fair comparison, all deep learning baselines utilize our dual-stream adaptive encoder as the backbone, only modifying the learning paradigm and final objective function.
\begin{itemize}[leftmargin=*] 
    \item \textbf{TrajClus}~\cite{lee2007trajectory}: A geometric unsupervised method employing trajectory partitioning and density-based clustering.
    \item \textbf{FixMatch}~\cite{sohn2020fixmatch}: A general-purpose SSL framework for comparison of core pseudo-labeling strategy.
    \item \textbf{DeepConvLSTM}~\cite{singh2020deep}: A widely adopted hybrid architecture for spatiotemporal sequence classification.
    \item \textbf{Transformer}~\cite{liang2022trajformer}: The standard transformer encoder for sequence-based modeling.
    \item \textbf{MultiScaleAttention}~\cite{jiang2020multi}: A specialized attention network designed to capture multi-scale temporal features in trajectories.
\end{itemize}

\subsection{Overall Performance}
Table \ref{tab:model_results_combined} presents the comparison across varying labeled ratios. \frameworkname{} consistently outperforms all baselines across all label regimes. In the label-scarce 5\% setting, our model (0.635) establishes a commanding 47.3\% lead over the best supervised method (DeepConvLSTM, 0.431), and significantly surpasses the unsupervised TrajClus (0.363). This dominance highlights the effectiveness of our framework in leveraging unlabeled data. Notably, \frameworkname{} at 5\% labels already exceeds the performance of all supervised models trained on 100\% of the data.
\begin{table}[!t]
\centering
\caption{Overall performance comparison on the TMI classification task (F1-Score). Best: \textbf{bold}; Second: \underline{Underline}.}
\label{tab:model_results_combined}
\begin{tabular}{l|cccc}
\hline
\textbf{Model} & \textbf{5\%} & \textbf{20\%} & \textbf{50\%} & \textbf{100\%} \\
\hline
TrajClus~\cite{lee2007trajectory} & 0.363 & 0.363 & 0.363 & 0.363 \\
DeepConvLSTM~\cite{singh2020deep} & 0.431 & 0.606 & 0.573 & 0.593 \\
MultiScaleAttn~\cite{jiang2020multi} & 0.430 & 0.427 & 0.594 & 0.580 \\
FixMatch~\cite{sohn2020fixmatch} & \underline{0.618} & \underline{0.700} & \underline{0.780} & \underline{0.769} \\
\textbf{\frameworkname{}} & \textbf{0.635} & \textbf{0.775} & \textbf{0.842} & \textbf{0.934} \\
\hline
\end{tabular}
\end{table}

When compared to the strong semi-supervised baseline FixMatch, \frameworkname{} maintains a 1.7\% gain in the 5\% setting, validating the efficiency of our graph-prototypical SSL strategy. More critically, while FixMatch suffers from performance saturation, plateauing around 0.769 F1 (likely due to accumulated pseudo-label noise), our model scales unimpeded to a peak F1 of 0.934 with 100\% labels. This 21.5\% performance gap between \frameworkname{} (0.934) and FixMatch (0.769) confirms that our design is not only a better SSL approach for sparse supervision but also a fundamentally superior architecture capable of exploiting the full supervisory signal for complex spatiotemporal tasks.

\subsection{Detailed Per-Class Analysis}
\begin{figure}[!t]
    \centering
    \includegraphics[width=0.3\textwidth]{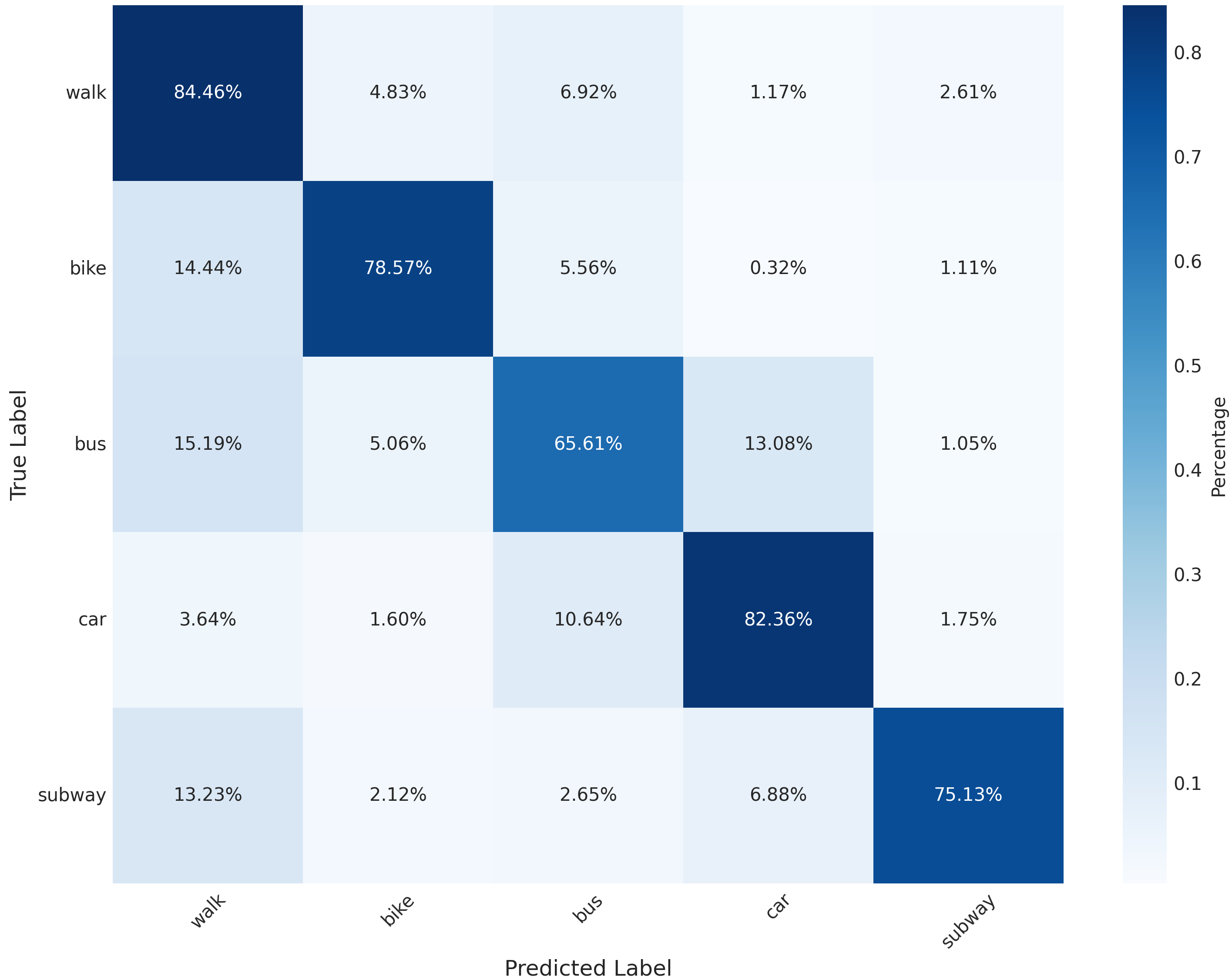} 
        \caption{Confusion Matrix of \frameworkname{}'s performance at the 20\% labeled data ratio.}
    \label{fig:confusion_matrix}
\end{figure}

A granular analysis of per-class performance is visualized in the confusion matrix (Fig. \ref{fig:confusion_matrix}) for the 20\% label rate. The matrix reveals that \frameworkname{} reliably classifies modes with distinct kinematic profiles. Strong diagonal values are observed for Car, Bike, and the minority class, Subway, suggesting the graph-prototypical core successfully extracts their unique motion signatures.

Conversely, the matrix shows that confusions primarily arise in modes with ambiguous or overlapping motion patterns. The Bus class, for instance, is frequently misclassified as Car (13.08\%) or Walk (15.19\%). This indicates difficulty in separating bus-specific stop-and-go road travel from continuous driving or slow, stop-like segments. Similarly, Walk segments show notable misclassification into Bike (14.44\%) and Bus (15.19\%), highlighting local ambiguities where slow walking activity is confused with cycling or vehicle dwelling. These mixed-mode confusions illustrate the intrinsic difficulty of TMI, which our manifold-aware approach addresses while providing transparent error diagnosis.

\section{Conclusion}
\label{sec:conclusion}
In this work, we addressed the critical challenge of  confirmation bias and failure to capture trajectory topological dependencies in Semi-Supervised TMI under extreme label scarcity. We proposed \frameworkname{}, a novel, context-free, graph-prototypical multi-objective SSL framework. Our core innovation lies in the synergistic use of dynamic graph regularization and prototypical anchoring to model the underlying transport network structure. This core is complemented by a robust, dual-filtered pseudo-labeling strategy that actively suppresses label noise, providing a primary defense against bias. Experiments on the GeoLife benchmark showed \frameworkname{} achieves state-of-the-art performance, with the model using only 5\% of labels outperforming all fully supervised baselines, demonstrating its ability to exploit intrinsic data structure and scale effectively with limited supervision. This confirms the high structural quality of our learned representations. Future work includes extending \frameworkname{} to online TMI for streaming data and exploring its applicability in other sequence-based tasks with limited labeled data.

\bibliographystyle{IEEEtran}
\bibliography{reference}

\end{document}